# Can ChatGPT Make Explanatory Inferences?
# Benchmarks for Abductive Reasoning

Paul Thagard, University of Waterloo
September, 2024
paulthagard.com, pthagard@uwaterloo.ca

**Abstract** Explanatory inference is the creation and evaluation of hypotheses that provide explanations, and is sometimes known as abduction or abductive inference. Generative AI is a new set of artificial intelligence models based on novel algorithms for generating text, images, and sounds. This paper proposes a set of benchmarks for assessing the ability of AI programs to perform explanatory inference, and uses them to determine the extent to which ChatGPT, a leading generative AI model, is capable of making explanatory inferences. Tests on the benchmarks reveal that ChatGPT performs creative and evaluative inferences in many domains, although it is limited to verbal and visual modalities. Claims that ChatGPT and similar models are incapable of explanation, understanding, causal reasoning, meaning, and creativity are rebutted.

When the Baltimore Harbor bridge collapsed in 2024, people keenly wanted to know *why* it collapsed. The immediate explanation was that the collapse resulted from collision with an enormous cargo ship, the Dali. But that conclusion raises many additional questions, such as why the Dali collided with the bridge's support structure, and why the collision was sufficient to take down the whole bridge. Similar questioning and inference are ubiquitous in numerous areas of human thinking, including science, medicine, technology, law, humanistic interpretation, and interpersonal relations.

I use the term *explanatory inference* to mean the creation and evaluation of hypotheses that explain puzzling occurrences. I will shortly clarify why I prefer this terminology to the usual terms "abduction" and "abductive inference". Explanatory inference is sometimes creative, producing novel hypotheses as often happens in scientific discovery. After 2 or more hypotheses have been proposed, we can evaluate them to determine which provides the best explanation of the available evidence. I will describe these two kinds of inference as *creative explanatory inference* and *evaluative explanatory inference.*

Explanatory inference has been used by people throughout recorded history, and more recently by computer programs that capture only a modicum of the capabilities of people. However, the 2020s have brought an explosion in the power of artificial intelligence to approximate human reasoning, through the use of generative AI techniques that include large language models. Models produced by leading companies include successive versions of ChatGPT from OpenAI, Claude from Anthropic, Gemini from Google, and Llama from Meta. Are they capable of creative and evaluative explanatory inference?

I use ChatGPT 4, one of the most advanced models available in April, 2024, to test the explanatory capabilities of current AI. I propose a set of benchmarks for explanatory inference that show that generative AI is astonishingly capable of a wide range of creative and evaluative explanatory inferences, although there are still some limitations that might be addressed by future research. Before laying out benchmarks for assessing their performance, I provide some historical background and terminological clarification.

## Abduction and Abductive Inference

More than a century ago, Charles Peirce coined the term "abduction" to mean both the initial formation of hypotheses and the assessment of their plausibility, distinguishing it from the much more familiar deduction and induction (Peirce 1992, Thagard 1977). Peirce's terminology varied, as his earlier writings used the term "hypothesis" for this kind of inference, and some of his later writings called it "retroduction". Subsequent writers have called such thinking "abductive inference", "abductive reasoning", or "abductive cognition".

I will avoid the terms "abduction" and "abductive" for several reasons. First, the terms are ambiguous between the creative and evaluative aspects of explanatory inference, which may require different mechanisms. Second, some recent writers have adulterated the term "abduction" to include non-explanatory forms of inference, such as coming up with mathematical axioms to yield theorems (Gabbay and Woods 2005, Magnani 2009). I am only concerned with inferences that contribute to explanations. Third, the use of "abduction" to mean inference rather than kidnapping is not familiar outside of narrow areas of philosophy and computer science. In contrast, the problems of creative and evaluative explanatory inference should be commonly comprehensible.

In philosophy, the leading writer on abductive inference since Peirce has been Lorenzo Magnani, who has developed a powerful *eco-cognitive* model of abduction which emphasizes that it is distributed, embodied, and multimodal in ways that my benchmarks acknowledge (Magnani 2001, 2009, 2017, 2022a, 2022b). I have the pleasure of dedicating this article to Lorenzo in appreciation of his excellent writings, and also of his organizational contributions to conferences and publications that have advanced understanding of abductive inference.

In philosophy, evaluative explanatory inference is widely known as *inference to the best explanation,* a term coined by Gilbert Harman, although the idea can be traced back at least to the renaissance (Blake, Ducasse, and Madden 1960; Harman 1965, 1973; Lipton 2004; Thagard 1978). Artificial intelligence researchers have largely used the term "abduction" for evaluative explanatory inference as it operates in fields such as medicine (Bylander, Allemang, Tanner, and Josephson 1991; Hobbs, Stickel, Appelt, and Martin 1993; Josephson and Josephson 1994; Peng and Reggia 1990). Psychologists sometimes emphasize the creative aspects of abductive inference, and sometimes its evaluative aspects (Haig 2005, Lombrozo 2012). This terminological variability across fields provides further reasons for my preference for spelling out the difference between creative and evaluative explanatory inference, while downplaying the term "abduction".

My theory and computational model of explanatory coherence is a contribution to evaluative explanatory inference (Thagard 1989, 1992, 2000, 2012). But I have also proposed computational accounts of creative explanatory inference and concept formation (Thagard 1988, 2012). The algorithms I have used to model creative and evaluative inference have been very different, but one surprising conclusion of the examination of ChatGPT is that the same computational processes may do both creation and evaluation. An even more surprising conclusion is that the performance of ChatGPT is superior to all previous computational models of creative and evaluative explanatory inference, including my own.

## Benchmarks for Explanatory Inference

In artificial intelligence, researchers commonly employ benchmarks for evaluating competing models in fields such as natural language processing and machine learning (Martínez-Plumed, Barredo, Ó hÉigeartaigh, and Hernández-Orallo 2021). I provide a novel set of benchmarks for assessing the extent to which generative AI models perform creative and explanatory inference. My assessment will not compare the models against each other, but rather compare ChatGPT 4 with human performance. This version, which appeared in 2023, is significantly more powerful than the free version ChatGPT 3.5 that has been available since November, 2022.

The new AI models are sometimes called "large language models", but that term is misleading because some models can operate with visual images and sounds as well as words. Another common term is "foundation models", which is obscure without saying what they are foundations of. I prefer the term "generative AI" for the general class of AI programs that are currently prominent. All of them are based on deep learning in neural networks, greatly enhanced by a method called "attention" that provides a powerful way of keeping track of context and relevance. The methods used are also called Transformers for obscure reasons.

The power of Transformers comes from a combination of factors (Vaswami et al. 2017). Vectors provide a mathematically rich way of representing data in many modalities, including language, vision, and sound. Vector processing, including the attention mechanism, can run in highly parallel fashion in modern computers using specialized processing units. Attention mechanisms weight the importance of different tokens in a sequence based on long-range dependencies for context and relevance. My 2021 book *Bots and Beasts* evaluated advanced AI models and estimated that human-like intelligence was at least decades away, but I was wrong (Thagard 2021). The capacity of ChatGPT to perform explanatory inference can be assessed by considering domains, modalities, hypothesis formation, and hypothesis evaluation.

**Domains**

Traditional AI models have been restricted to narrow domains such as medicine or game playing. Because generative AI models can be trained on billions of documents and Web sites, they cover a mammoth proportion of human knowledge. They should therefore be able to perform explanatory inference in countless domains such as the following.

*Science* In fields such as physics, chemistry, biology, and the social sciences, scientists generate hypotheses to explain their experimental findings. They also evaluate these hypotheses to determine which theories provide the best explanation of these findings. Can ChatGPT 4 create and evaluate hypotheses in all these domains?

*Medicine* Explanatory inference is important to medicine as a branch of applied science, in the development and assessment of theories about the causes and treatments of disease. But medicine is also the practice of helping individuals by diagnosing the causes of their symptoms. Diagnosis requires creative inference in guessing what disease or diseases might explain a patient's symptoms, and also evaluative inference in deciding which of the possible diseases best explain the symptoms.

*Law* The legal system requires explanatory inference for both the detective work of determining who might have committed a crime, and also for the trial work of determining whether the accused is actually guilty. Detective work is primarily creative explanatory inference to identify suspects, although it can also be evaluative if there are multiple plausible suspects. Trial work is largely evaluative explanatory inference, with the prosecution aiming to convince the jury that the suspect is guilty, whereas the defense tries to show that the evidence is not sufficient to justify guilt beyond a reasonable doubt. Sometimes the defense makes its case by suggesting alternative suspects who might have committed the crime, i.e. alternative hypotheses.

*Technology* Failures of technology such as the Baltimore bridge collapse require both the creation of hypotheses to explain what went wrong and the evaluation of the resulting competitors. One illustration is debugging of computer code to determine why it does not perform as desired. But creative and evaluative explanatory inference are also relevant to the development of technology, as designers form and assess hypotheses about what might work to solve outstanding problems such as designing machines and algorithms. Technology covers myriad everyday practices such as cooking, gardening, carpentry, and car repair, all of which require hypothesis formation and evaluation.

*Artistic and Literary Interpretations* Humanistic scholars and ordinary people can ask questions about the deep meaning of works of art such as a van Gogh painting, an Emily Dickinson poem, or a Bob Dylan song. Such interpretations require the generation and evaluation of hypotheses that could explain why the artist produced their art and also what it means in a broader historical and social context.

*Interpersonal Relations* Interactions between people require a steady flow of explanatory inferences concerning the mental states of others, including beliefs, motives, and emotions. People naturally create hypotheses that would explain why people are behaving in unexpected ways, although they are often lazy about evaluating the plausibility of these hypotheses.

People vary dramatically in the extent to which they are experts in these domains, and many people are poor at explanatory inferences in areas where they lack the required knowledge, such as my limitations in car repair. But ChatGPT should be able to create and evaluate explanatory hypotheses in all of these areas.

**Modalities**

Hypotheses and the evidence they explain are typically expressed in words, as in the sentence "The Dali collision explains the collapse of the Baltimore Bridge." But hypotheses can also be expressed in other representational formats such as pictures, sounds, and smells. In cognitive science, the term "modality" applies to all the different modes of representation of which humans are capable, including visual (a picture or mental image of the ship hitting the bridge), auditory (the sound of the bridge crashing), olfactory (the smell of spilled oil), and so on. How does ChatGPT fare with the full range of modalities found in human cognition? For each modality, we can ask whether a system such as a human or a computer can use that modality for inputs such as the evidence to be explained, and for outputs such as the hypotheses that do the explaining.

*Verbal* Sentences and words can represent evidence, hypotheses, and the explanatory relations among them. Humans and ChatGPT are adept at generating and manipulating verbal representations, but non-human animals and older computer programs are not.

*Visual* Most people are skilled at recognizing and imagining pictures. They take visual information such as a picture of a bridge crash as inputs to explanatory inference. They can also produce outputs in the form of visual hypotheses such as a static picture of a ship hitting a bridge, or even better a mental movie in which the ship approaches the bridge and crashes into it. Visual explanatory inferences are largely creative, but they can also be evaluative if pictures and movies are assessed to see which provides the best explanation of the evidence.

*Other External Senses.* People can also use sound, smell, taste, and touch as perceptual evidence to be explained. Explanatory inference can be prompted by a large bang, a foul smell, a noxious taste, or an irritating touch. Less obvious is whether these modalities can also be used to represent explanatory hypotheses, but there are at least a few cases such as when we imagine that someone has tasted a foul cheese that is spat out. A fart in an elevator can be both a fact to be explained and a hypothesis to explain why people move away from the perpetrator.

*Internal Senses* Human thinking also employs representations resulting from internal senses, including pain, hunger, heat, cold, kinesthesia, proprioception, balance, and sexual arousal. These senses can inspire explanatory inference, for example the generation of explanations of why one's stomach hurts. But they can sometimes also be explanatory, for example when I remember what it is like when I have stomach ache in order to explain someone else's pain behavior such as moaning and complaining.

*Emotions* People have hundreds of different emotions such as happiness, sadness, fear, anger, surprise, disgust, shame, and gloating. Feeling such emotions can spark creative and evaluative explanatory inferences, such as figuring out that someone is sad because of a work disappointment. Emotions can also be explanatory, both through verbal representations such as "They are crying because they are sad" and through experienced emotions derived through empathy. Pat can explain Sam's behavior by imagining being in Sam's situation, feeling an emotion, and transferring that feeling to Sam. Humans can use all of these modalities in explanatory inferences, but ChatGPT is much more limited.

**Hypothesis Creation**

Human creative explanatory inference uses various methods of hypothesis formation, including reasoning that is causal, existential, analogical, manipulative, based on conceptual combination, augmentative, distributed, and initiated by emotions (Thagard 1988, 2006; Magnani 2009). How many of these can ChatGPT perform?

*Causal* The main method of creative explanatory inference about a puzzling occurrence is conjecturing possible causes, as in the bridge example. Imagining that the ship hit the bridge with a huge amount of force would provide a causal explanation of the collapse, so we form the hypothesis that that is what happened.

*Existential* Sometimes explanation requires postulating the existence of something not previously observed, for example guessing that a perturbation in the motion of a planet is the result of the existence of an undiscovered planet. Inferring that someone has a mental state such as a subtle emotion is also existential creative explanatory inference.

*Analogical* Analogy can be a useful tool for generating explanatory hypotheses (Thagard 1988, Holyoak and Thagard 1995). Darwin formed many of his explanations concerning natural selection by analogy with the familiar process of artificial selection by breeders.

*Manipulative* Hypothesis formation can be enhanced by working with external models such as physical devices or diagrams (Magnani 2017). For example, someone trying to figure out what forces might operate in a bridge collision could draw a diagram with arrows representing causal relations.

*Conceptual Combination* Peirce said that abduction is the only kind of inference that introduces new ideas, but did not describe how this works. The common sentential characterization of abduction is: Hypothesis H would explain evidence E. Evidence E. So maybe H. This schema says nothing about how H is created. However, new hypotheses can be formed by combining previously existing concepts, for example when Darwin formed *natural selection* from previous concepts *natural* and *selection.* Conceptual combination can be modeled both by symbolic psychological processes and by neural processes (Thagard 1988, 2012).

*Augmenting Explanations* Sometimes new hypotheses are formed through the attempt to fill in the gaps in available explanations. Inheritance was a crucial part of Darwin's theory of natural selection, but he had no good theory of inheritance which was eventually filled in by genetic theory. Mechanist explanations often have gaps that can be filled in by sub-mechanisms.

*Distributed Creation* Hypotheses are sometimes generated by people working together who can be more productive than individuals.

*Emotional Initiation.* Why do people set out to form new explanatory hypotheses? Peirce recognized that new hypotheses can be prompted by emotions such as surprise and doubt, and other emotions can similarly motivate creative explanatory inference. The desire to explain can come from emotions such as wonder, curiosity, and puzzlement, but also from less noble emotions such as ambition, greed, envy, and anger.

To match human performance in creative explanatory inference, ChatGPT would have to display all 7 aspects of hypothesis formation. We will see that it can accomplish at least 5 of these, and has good prospects of accomplishing another.

**Hypothesis Evaluation**

Hypothesis formation is a highly dangerous form of inference, as people often generate falsehoods, for example in the form of conspiracy theories (Thagard 2024). Within days of its occurrence, the Baltimore bridge collapse was being tied to spurious causes such as foreign interference and covid vaccinations. Evaluation of proposed hypotheses is therefore crucial before they can be accepted as likely to be true. Proposed criteria for evaluation include explanatory breadth, being explained, simplicity, analogy, prediction, and probability. How well does ChatGPT apply such criteria?

*Explanatory Breadth* Prefer a hypothesis that explains more than its competitors.

*Being Explained* Prefer a hypothesis that is itself explained by a deeper mechanism.

*Simplicity* Prefer a theory that uses fewer hypotheses, makes fewer assumptions, proposes fewer entities, or is otherwise less complex than its alternatives.

*Analogy* Prefer a theory whose explanations are similar in causal structure to accepted explanations.

*Prediction* Prefer a theory that makes testable predictions that could falsify it.

*Probability* Prefer the more probable hypothesis as calculated by Bayes' theorem.

**Conceptual Issues**

Explanatory inference in humans presupposes crucial concepts such as explanation, understanding, causality, meaning, and creativity. Critics of generative AI models have claimed that they are only "stochastic parrots", using statistical patterns in the data to give the illusion of intelligence (Bender, Gebru, McMillan-Major, and Schmitchell 2021). For ChatGPT to count as performing explanatory inference, both creative and evaluative, it needs to have some grasp of these concepts, even though it gets its performance results solely from training on data rather than on interactions with the world.

*Explanation* Does ChatGPT explain anything when it generates texts and images just by predicting the next word?

*Understanding* Is ChatGPT capable of any understanding when it lacks world-related semantics, conscious experience, and emotional concern about anything?

*Causality* How can ChatGPT comprehend causality when it does not interact with the world and has no body to indicate the difference between a push and a pull, or even between causality and correlation?

*Meaning* Are the sentences generated by ChatGPT meaningful when its words lack connections to the world?

*Creativity* Is ChatGPT only mimicking creativity by generating hypotheses, essays, stories, poems, jokes, lyrics, music, and computer code, when it lacks emotional responses to determine if these are results are valuable and hence creative?

These are difficult questions that must be addressed before we can conclude that ChatGPT and other generative AI model are capable of explanatory inference.

**Tests of Explanatory Inference by ChatGPT**

I now report systematic tests of the capabilities of ChatGPT 4 to generate and evaluate explanatory hypotheses. This program is available with a paid subscription at openai.com and is significantly more powerful than its predecessor ChatGPT 3.5 (Bubeck 2023, OpenAI 2023). The version I used was trained in April 2023, and may soon be superseded by ChatGPT 5. For lack of space, I will not quote the texts generated by ChatGPT, but files containing these texts are available on request. Better, you can run the tests yourself. Let us now consider the performance of ChatGPT with respect to domains, modalities, hypothesis creation, hypothesis evaluation, and conceptual issues.

**Domains**

I tested ChatGPT's performance in more than 20 domains by giving it these prompts, where X is something that requires explanation.

- Evaluate competing hypotheses about X.
- Generate a novel hypothesis about X.

The domains tested included examples from physics, chemistry, biology, geology, climatology, neuroscience, psychology, sociology, economics, medicine (both theoretical and applied), geoengineering, law, car repair, computer repair, art interpretation, poetry interpretation, and literary history. You can easily find appropriate topics for explanatory hypotheses in a domain D by this prompt:

- What are controversial issues in D?

ChatGPT's performance in all of these domains was stellar, roughly at the level of a sophisticated graduate student and definitely superior to an average undergraduate. In the domains I know well, I can attest that its answers were consistently comprehensive and accurate. It always considered at least four alternative hypotheses, evaluated them with respect to what they can and cannot explain, and reached a reasonable overall conclusion.

For example, when asked to evaluate competing hypotheses for global warming, ChatGPT considered greenhouse gas emissions, solar variability, volcanic activity, natural climate variability, and orbital changes. It concluded with the scientifically accepted view that greenhouse gas emissions are the primary driver of current global warming. When prompted, ChatGPT also generated the hypothesis that global warming results from "microbial methane regulation disruption". If the ability to evaluate and generate hypotheses across more than 20 fields is not impressive enough, ChatGPT can also provide its responses in languages that include Spanish, French, German, Italian, Portuguese, Dutch, Russian, Chinese, Japanese, and Korean.

ChatGPT's amazing ability to produce intelligent responses in so many domains is the result of its being trained on billions of electronic documents available through the Web and other sources. No human could possibly have been exposed to so much information. With respect to explanatory inference, the program has not only passed the Turing test of being able to imitate human responses, it far surpasses any human in its breadth of knowledge.

When asked how it came up with the hypotheses that it generated, ChatGPT claims that it uses interdisciplinary knowledge across various domains to create novel scenarios and domains, while exploring their implications. I have no idea whether this account is accurate, just as I have no idea whether ChatGPT's account of how it composes its poems by analysis and metaphor is right. After all, ChatGPT's central function is just to produce the next word! Below in the conclusion section I consider the possibility that ChatGPT's learning enables it to acquire complex cognitive skills, which might include generating hypotheses and writing poetry. Similarly, human brains are not explicitly trained to produce explanatory hypotheses or works of art, but they have these skills as side effects of simpler mechanisms. This cascade of cognitive complexity operates by repeated emergence of productive skills, right up to consciousness.

ChatGPT acknowledges its limitations with respect to: real-time information only available since its last training, advanced expertise not available in documents, proprietary and confidential information, local knowledge in specific cultures, and multimedia content. These limitations are found in humans as well, so they do not undercut my judgment that ChatGPT is superior to any human in the range of domains for explanatory inference. I asked Google search for some obscure branches of science, and the result included pedology (soil science). Then I asked ChatGPT to evaluate competing hypotheses in pedology and it coherently compared hypotheses on several topics including pedogenesis (soil formation), for which it also generated a novel hypothesis. Most human expertise is restricted to a few domains, but I have yet to find a domain in which ChatGPT cannot do explanatory inferences.

**Modalities**

ChatGPT is enormously adept at working with words and sentences in English and other languages. But how does it fare with other modalities such as vision, sound, internal sensations, and emotions? The currently best-known examples of generative AI are large language models that manipulate words, but some AI models can also work with pictures and sounds.

Unlike its predecessors, ChatGPT 4, can take visual inputs and produce visual outputs, including ones that are explanatory. I have given ChatGPT pictures of bridge collapses and car crashes which it competently interprets in words. For example, ChatGPT interprets the scene in figure 1 as a car accident, and even makes the verbal explanatory inference that it resulted from a head-on collision.

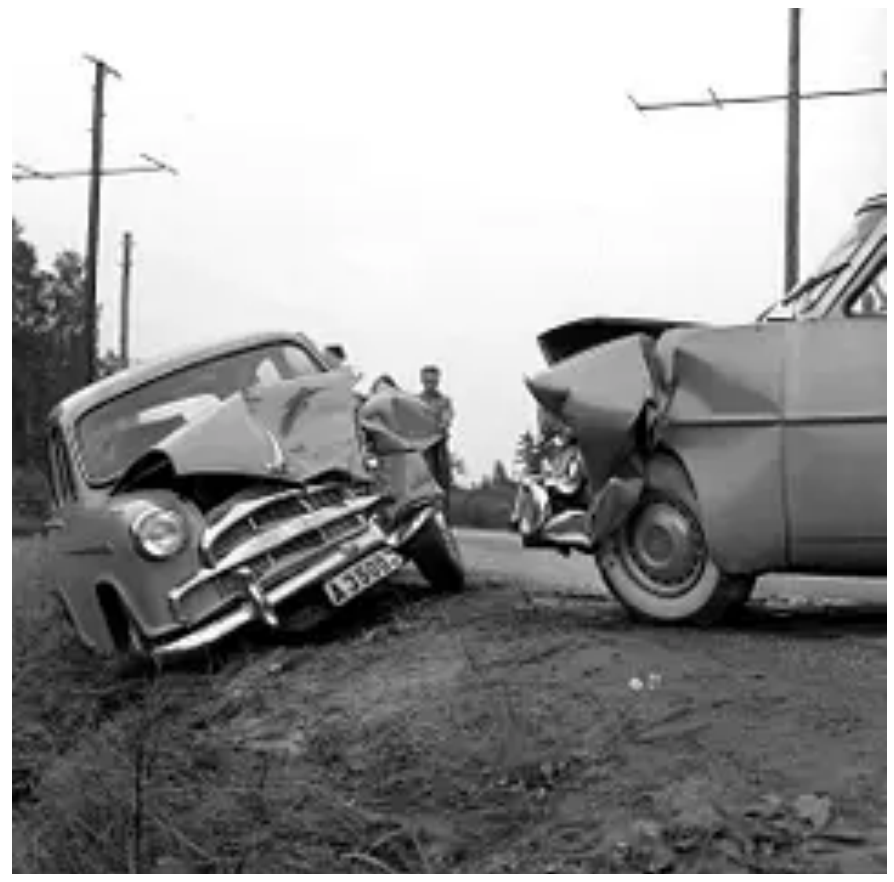

**Figure 1** Source: https://picryl.com/topics/car+crash, public domain.

The visual hypothesis of such a collision is generated by ChatGPT using the image generator DALL-E as shown in figure 2. The computation here does not seem to be directly from visual evidence to visual explanation as humans probably do, but rather from visual description to verbal explanation to visual depiction. Nevertheless, ChatGPT is clearly operating in the modality of vision.

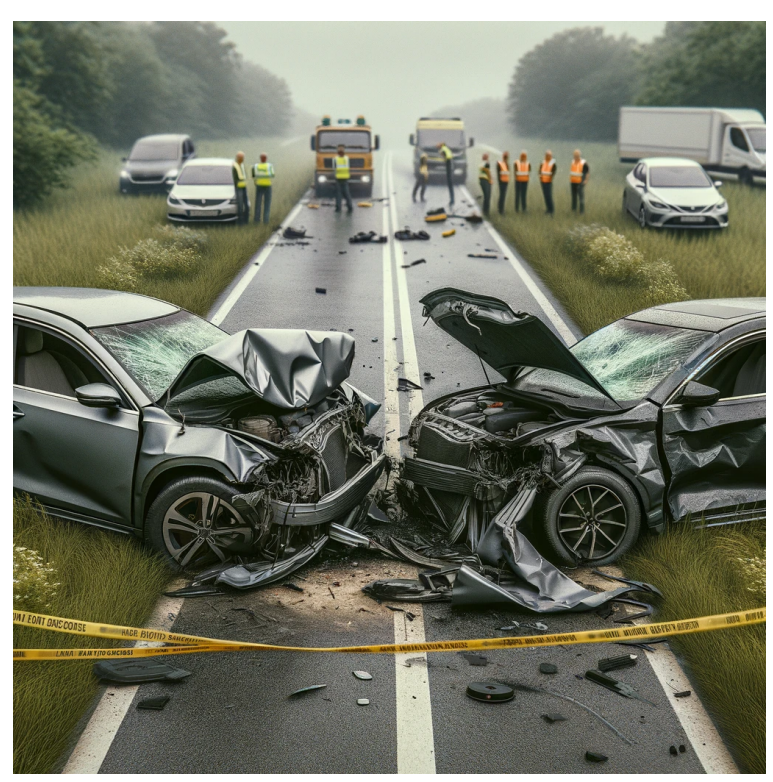

**Figure 2** Visual explanation of a car crash, produced by ChatGPT using DALL-E technology.

ChatGPT is also capable of generating an image to explain how a Tesla got a scratch on its door, as in figure 3. Note that this image performed hypothesis generation, because the event of the bicycle crashing into the car was not part of its instructions.

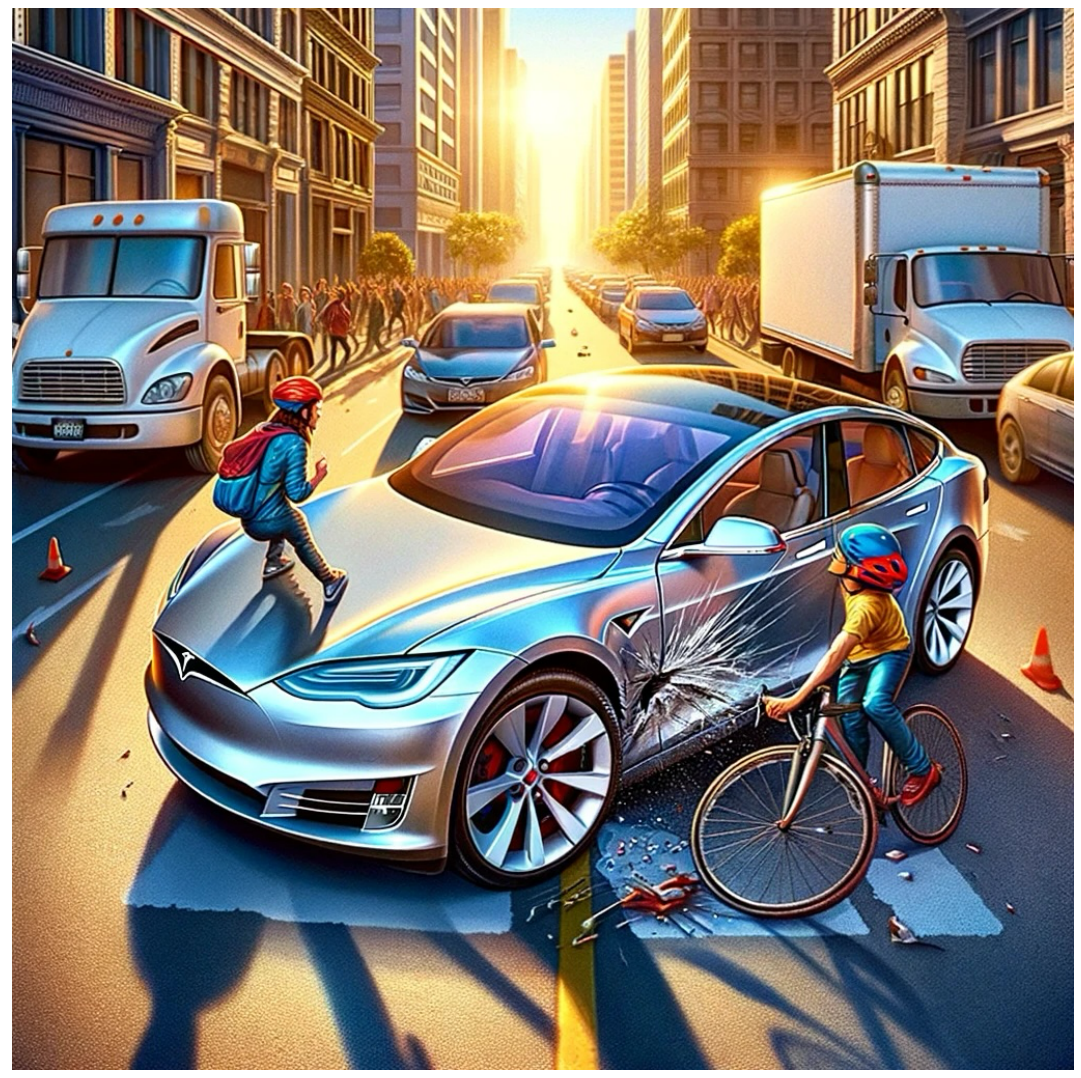

**Figure 3** Image of a Tesla getting scratched, produced by ChatGPT 4.

ChatGPT also produces a visual representation of various possible causes of a bridge collapse, as shown in Figure 4. Pictures are excellent at hypothesis generation, but play a lesser role in hypothesis evaluation.

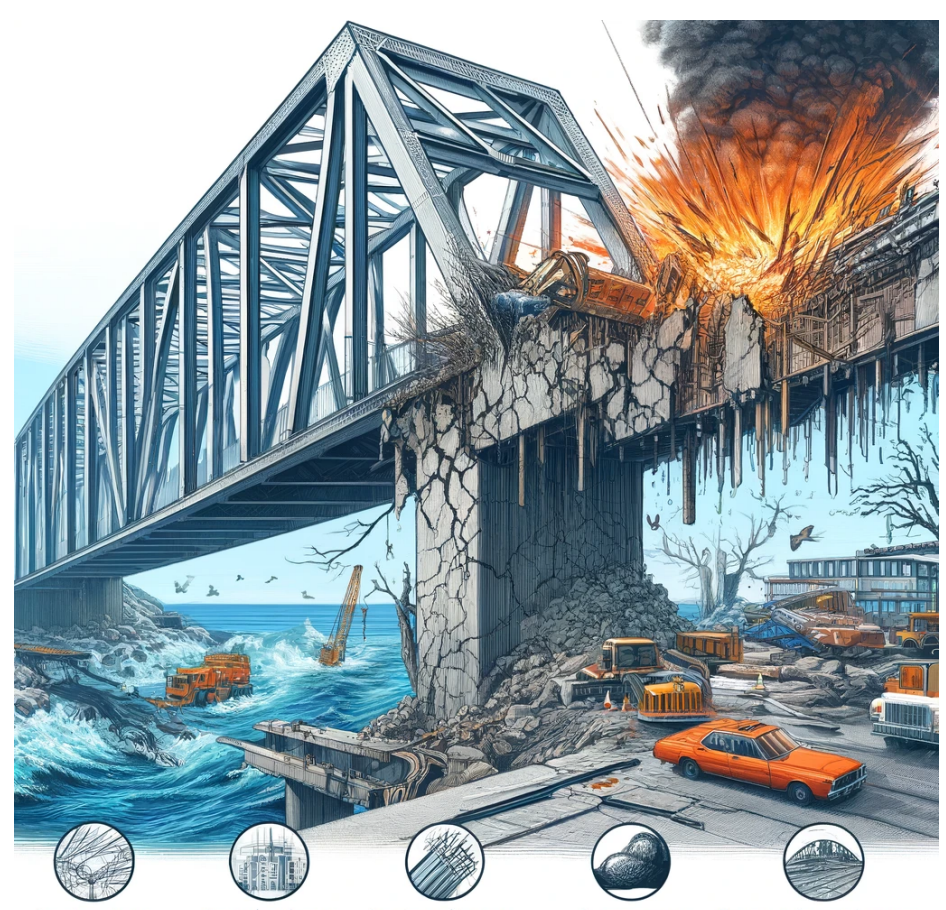

**Figure 4** Visual explanations of bridge collapse, produced by ChatGPT 4.

ChatGPT only produces static pictures, and more vivid explanations would be produced by movies, just as people can create dynamic images in their heads. OpenAI has developed a program called Sora that is capable of taking verbal instructions and generating impressively realistic movies (OpenAI 2024). However, it is still experimental and not available for general use. Gemini and Llama also work with visual representations.

ChatGPT's performance with visual modalities is inferior to human visual reasoning. Although it can take visual inputs and produce explanatory outputs, it lacks the flexibility of humans to produce dynamic visual images. Perhaps that will be incorporated in later versions. Nevertheless, ChatGPT does enough with visual representations that we can judge it to be multimodal.

For humans, the main role of auditory representations concerns surprising noises that require explanations, such as bangs, backfires, gunshots, moans, and off-tune music. Less commonly, we can also form auditory hypotheses, such as imagining a loud boom that might have made someone jump. ChatGPT does not take sounds as inputs or outputs, so it is incapable of auditory explanations.

However, other generative AI programs can be trained on auditory inputs and produce novel outputs such as songs. OpenAI has a program called Jukebox that produces music in a wide range of musical styles such as rock in the style of Elvis Presley, with original music accompanied by realistic voices and appropriate lyrics (https://openai.com/research/jukebox), I suspect it would not be hard to merge this technology with ChatGPT in order to allow the generation of auditory hypotheses, or at least the generation of hypotheses that explain input sounds. Because all of ChatGPT's inferences involve the manipulation of vectors, and sounds are easily translated into vectors, ChatGPT should be eventually capable of auditory explanatory inference. Google reports that Gemini works with sound and movies (Gemini Team Google 2024).

ChatGPT is incapable of olfactory explanations, but technology is expanding to enable computer representations of smells (Lee et al. 2023). Similarly, robotic sensors are already capable of encoding touches and tastes to be translated into vectors that train generative AI models for explanatory inferences. ChatGPT is already capable of explanatory inference using visual representations, and extensions to sound, smell, taste, and touch are highly feasible. So multimodal inference for these senses is within the capabilities of ChatGPT.

In contrast, the prospects for explanatory inference concerning internal senses and emotions are far more problematic. Consider the concept of *hangry* which applies to people who are angry because they are hungry, for example someone who skipped breakfast and then yells at other people. Hunger is a complex sense that involves brain areas such as the arcuate nucleus in the hypothalamus responding to bodily signals from pangs in the stomach, body weakness, and head lightness (Beutler et al. 2017). No technology exists for machine recognition of these hunger signals, so they cannot be used to train ChatGPT or similar models. Perhaps physiological understanding of hunger will increase sufficiently to allow simulations in vectors of all the inputs to the hippocampus, but this appears to be far off. Moreover, people's feelings of hunger also depend on broader coherence considerations such as duration of lack of food, levels of key hormones ghrelin and leptin, visual and olfactory presentation of attractive food, and social context (Thagard forthcoming).

Emotion is even more complex, because it requires the brain to integrate neural firings that represent the situations that the emotion is about, physiological changes such as breathing, heart, and adrenaline, and cognitive appraisals of the situation (Thagard,

Larocque, and Kajić 2023). Although computational models of such integration have been developed, not nearly enough is known to translate inputs and outputs of emotions into vectors that could train a future ChatGPT. For example, anger is typically about situations such as nasty people, is associated with physiological changes such as increased heartbeat and cortisol levels, and is connected to appraisals that someone or something is blocking the accomplishment of one's goals. But putting all this into a ChatGPT model will require major breakthroughs in the understanding of biological processes. ChatGPT can give a lucid and comprehensive description of the physiological, psychological, and social factors that go into explanations based on people being hangry. But it is totally ignorant of the sensory and emotional aspects of hangry explanations.

Some AI researchers are already claiming that the new models are capable of kindness and empathy, but the models are only currently capable of faking such emotional reactions. Empathy requires imagining yourself in someone else's situation and using your past experience to imagine how they are feeling (Thagard 2019). You can empathize with someone being hangry by remembering a situation where you got grumpy with people because of lack of food, and supposing that the other person is feeling something like what you previously felt. Empathic explanations are based on nonverbal analogies not currently available to computational models. Although ChatGPT does not have emotions or empathy, it has excellent verbal comprehension of them through its training data, and does well at discussing and even faking emotions and empathy.

**Hypothesis Formation**

Creative explanatory inference consists of forming hypotheses that might explain puzzling facts. The examples that ChatGPT produced as part of my domain investigations shows that it can generate hypotheses in more than 20 different fields. But we need to examine more closely whether it can employ specific aspects of human hypothesis formation. Tests show that ChatGPT is excellent at forming hypotheses based on causal reasoning, existence postulation, analogy, conceptual combination, and mechanism augmentation. Future versions are likely to be capable of manipulative hypothesis formation using external interventions, and distributed hypothesis formation involving multiple agents. But because the new AI models lack emotions, they are incapable of forming hypotheses driven by emotions such as surprise and doubt.

The most common form of creative explanatory inference is causal reasoning with the structure: A causes B; B; so maybe A. ChatGPT astutely distinguishes this form of reasoning from the fallacy of affirming the consequent: If A then B; B; therefore A. It gives the example: Lightning causes thunder; we hear thunder; so maybe there was lightning. All of the 20 domain examples I gave can be construed as causal reasoning, on the plausible assumption that the task of explaining something usually means providing a cause.

For example, I asked ChaGPT what might be the cause of Pat having a runny nose, cough, and fever, and it suggested 6 illnesses such as influenza as possible hypotheses. This interpretation assumes that ChatGPT has some understanding of causality, an issue I address below. I asked ChatGPT how it generates causal hypotheses and it gave a long response that attributed its ability to a combination of pattern recognition, language modeling, contextual understanding, and heuristics such as temporal ordering, correlation, and domain-specific causal relations such as between diseases and symptoms. Causal reasoning seems to be another emergent skill that ChatGPT gains from its training data.

One of the most powerful forms of creative explanatory inference is the generation of hypotheses about the existence of non-observable entities. ChatGPT does what I once called "existential abduction" with aplomb. Given a problem of explaining the perturbation of a planet, it suggests there might be another, unknown planet that is responsible. When I asked it to generate a novel hypothesis about an unknown sub-atomic particle, ChatGPT proposed the existence of a new particle called a "chrono-particle" that plays a role in time perception. This hypothesis is not very good, because time cells and other neural mechanisms provide a much better explanation of people's sense of time. But the purpose of hypothesis formation is not to get explanations immediately right, just to put some new ideas on the table for subsequent assessment.

In a similar vein, I asked ChatGPT to generate a novel deity that would explain global warming. It proposed Thermara, goddess of elemental balance, whose distress causes environmental changes. Religious explanations are deficient compared to scientific ones (Thagard 2010); but this proposal is definitely original, as on the Web Thermara is a towel warmer rather than a god. It might seem that ChatGPT has no scruples about generating hypotheses, but it balked when I asked it to generate a new hypothesis to explain why vaccines cause autism. Presumably as the result of human-guided reinforcement learning, ChatGPT declined the task on the grounds that scientific evidence does not support a link between vaccines and autism.

Use of analogy has been well documented as a source of novel explanations, and ChatGPT has an impressive ability to work with analogies and metaphors. Some of this success is the result of having been trained on material that used the analogies, but I have also frequently found it does a good job of elucidating analogies that it has not likely seen before. In April, 2024, I came across a comparison of the problem of AI systems being trained on data produced by AI with the inbreeding problem of the Habsburgs who had many birth defects. In this an many other cases, ChatGPT quickly lays out the appropriate correspondences, and it can even identify problems with the analogy in the form of mismatches. Keith Holyoak and his colleagues have systematically evaluated the performance of ChatGPT on analogies and metaphors, finding that the ChatGPT 4 version is particularly strong (Ichien, Stamenković, and Holyoak 2024; Webb, Holyoak, and Lu 2023).

ChatGPT is also capable of generating analogies on command, for example when asked to provide 4 analogies for conscious experience: a stream, a theater, a kaleidoscope, and a symphony orchestra. These could easily be looked up from texts used in training, so I asked ChatGPT to produce a novel analogy that generates a novel hypothesis that explains consciousness. I have written a book on consciousness and have thoroughly read the relevant literature, but ChatGPT came up with something I had not seen. It proposed that consciousness is analogous to an ecological system with consciousness resulting from "cognitive biodiversity". Moreover, ChatGPT spelled out the analogical explanation by going into detail about interconnectedness, adaptation, evolution, niche specialization, and emergence. It even generated predictions that could be used to test the hypothesis! In the end, I think this hypothesis is inferior to my own account of consciousness as resulting from four brain mechanisms, but I am still impressed by ChatGPT's expansive use of an unusual analogy. Perhaps, it picked up the idea of connections between consciousness and ecology from publications, but its development of the analogy strikes me as original (compare Goerner and Combs 1998, Mudrik, Hirschorn, and Korisky 2024).

The simple schema *A causes B, B, therefore maybe* A underestimates the potential creativity of explanatory inference, because it assumes that A has already been formulated. New concepts can be generated by computational procedures for conceptual combination, where new concepts construed as verbal data structures or patterns of neural firing are combined into new ones. Can ChatGPT form new concepts by conceptual combination?

The novel hypotheses generated by ChatGPT across multiple domains all involve original conceptual combinations, including: dimensional oscillation, quantum biogenesis, quantum neuro-resonance, environmental stress-induced viral evolution, microbial methane, regulation disruption, collective genius, quantum entanglement bonding, perceptual quantum coherence, cognitive elasticity, network cascade, and galactic information exchange. Whether these novel combinations are valuable is a separate question, to be answered by whether the theories they inhabit are better explanations than alternatives.

These examples show that ChatGPT is excellent at explanatory conceptual combination, as indicated by the generation of novel phrases. ChatGPT does conceptual combination very differently from previous computational models, using its usual methods of pattern recognition, contextual understanding, and creative synthesis. Randomness plays little role in its generation of sentences and their concepts.

I noticed another kind of hypothesis formation occurring in the attempts to explain the origins of COVID-19. The two main theories are a natural zoonotic spillover event and a laboratory-associated incident. Both theories have major gaps such as lack of knowledge of the specific animal or the specific lab accident that might have been responsible. Another example is the problem of explaining consciousness, where many neurological and psychological causes have been identified but there remains a gap in their explanation of experience. The process of filling in explanatory gaps by forming in hypotheses might be called "augmentative abduction" because it augments currently available explanations. ChatGPT performs well at generating potential intervening causes in both the COVID-19 case and the consciousness case. For COVID-19, ChatGPT speculates that the virus might have jumped from bats to another host and then to people, mutating along the way. This possibility augments the zoonotic hypothesis by filling in gaps in the causal chain from animals to human disease, generating topics for further research.

What about Magnani's idea of manipulative abduction which concerns developing explanations using external physical methods such as drawing diagrams? ChatGPT is currently incapable of such external actions, but research is active in finding ways for generative AI models to control robots using readily available programming tools (Metz 2024). The future could bring a ChatGPT version that uses external manipulation in its explanatory inferences.

Magnani has emphasized that human cognition is not just one mind acting alone, but is distributed across multiple people and their environments. Creative explanatory inference is often distributed, as in the collaboration of Francis Crick and James Watson in discovering the structure of DNA, which also included their interactions with other researchers and use of physical models of molecular structure. The only way that ChatGPT uses distributed cognition currently is that it interacts with individual people who prompt it to perform various tasks. In the future, ChatGPT might become socially distributed by interacting with other generative AI programs like Claude, or by working more collaboratively with groups of people. Such expansions are already feasible given current technologies, for example by the development of APIs (advanced programming interfaces) that connect ChatGPT with other programs. Hundreds of such mini-GPTs have already been developed that connect ChatGPT with commercial programs such as Expedia.

I conclude with the one aspect of creative explanatory inference that seems totally beyond the range of ChatGPT. Human hypothesis formation is often spurred by emotional reactions such as surprise. Without emotions, ChatGPT cannot have its hypothesis generation produced by the emotions that drive humans to be creative, such as doubt, surprise, wonder, puzzlement, curiosity, ambition, greed, and envy. I asked ChatGPT how each of these emotions can lead to hypothesis formation and to give examples from the history of science. For every emotion, the answer was both psychologically plausible and historically accurate! Hence although ChatGPT does not use emotions to form hypotheses, it seems to understand as well as any psychologist or historian how this process works. It also expanded my list of emotional drivers of creative hypothesis formation to include awe, frustration, dissatisfaction, confusion, competitiveness, fear, anxiety, optimism, and hope, accompanied by plausible examples.

ChatGPT can not only generate hypotheses, it can also help in the scientific investigations that surround specific formations. Here are some prompts that produce illuminating results concerning a topic X.

- What are controversial issues about X?
- Generate important questions about X.
- Generate experiments relevant to X.

I have used a method I call "chatstorming" which proceeds by getting ChatGPT to generate questions, provide answers to the questions, and then generate follow-up questions and answers. I used this method productively in writing a chapter on musical consciousness (Thagard 2023; Thagard forthcoming, ch. 4).

In sum, ChatGPT performs well on many aspects of creative explanatory inference, forming hypotheses using causal reasoning, postulating existence, analogies, conceptual combination, and augmentation. It could probably be expanded to become manipulative and socially distributed. But lack of human bodies prevents ChatGPT from having emotional inspirations for generating explanatory hypotheses, and its motivation to do so is entirely derivative from human instruction. Overall, however, we should be impressed by ChatGPT's ability to produce hypotheses that provide new ideas.

**Hypothesis Evaluation**

Generating hypotheses is indispensable to human cognition, but is a highly risky business, as we see from frequent failures in science and everyday life. The risk is enormously reduced by not accepting a hypothesis unless it qualifies as the best explanation of the available evidence, based on comparison with competing hypotheses. How good is ChatGPT at evaluative explanatory inference?

The many domain examples mentioned above show that ChatGPT is at least competent at inference to the best explanation. Asked to evaluate competing hypotheses about X, it compares various alternatives with respect to what they can and cannot explain. It picks a winner, which is reasonable in the fields in which I have enough expertise to judge. The question remains, however, whether ChatGPT is capable of evaluating competing hypotheses with respect to the relevant epistemic standards that include explanatory breadth, simplicity, being explained, analogy, testability, and probability.

My tests of the ability of ChatGPT to do evaluative explanatory coherence begin with this simple prompt: If one hypothesis H1 explains 3 facts, and another H2 explains 2 facts, which is better? ChatGPT surprised me by responding that it would be "oversimplifying" to choose a hypothesis based only on the number of facts explained. It reasonably suggested that the evaluation should also take into account the relevance and significance of the facts explained, the quality of the explanation with respect to depth and coherence, predictive power concerning future observations, simplicity in the sense of making fewer assumptions, consistency with established theories, falsifiability, and scope as part of a broader theoretical framework. These are good considerations that expose the limits of my intentionally naïve question, except for falsifiability which fails as a standard for evaluating theories because it is either too strict or too slack (Thagard 1988). Further prompts are required to determine what ChatGPT means by these factors, so for each I asked it why the factor is important for theory evaluation as illustrated by examples from the history of science.

ChatGPT did not do a great job of elucidating relevance and significance beyond saying that the facts explained should be within the theory's domain and pertinent to distinguishing it from competing theories. But the 4 historical examples it provided were on target: heliocentric vs. geocentric theories in astronomy, Newtonian mechanics vs. Einstein's theory of relativity, spontaneous generation vs. the germ theory of disease, and Lamarckism vs. Darwinian evolution. For each example, ChatGPT provided an accurate review of the controversy, description of some key facts to be explained, and discussion of which facts are significant in being explained by one theory and not the other. Despite my 5 decades of experience in the philosophy and history of science, I could not have done better.

ChatGPT did not add much about how the quality, depth, and coherence of explanations are important for theory evaluation, but applied them reasonably to 4 historical examples: atomic theory in chemistry, theory of evolution by natural selection, general relativity, and germ theory of disease. I think that quality of explanation can be better clarified in the context of the view that explanations often describe mechanisms, which are combinations of interconnected parts whose interactions produce regular changes (Thagard 2019, Craver and Darden 2013). Moreover, ChatGPT's characterizations of depth and coherence are superficial compared to my rigorous accounts of depth as explanation by lower-level mechanisms and coherence as computation of satisfaction of multiple constraints (Thagard 2012). Nevertheless, ChatGPT's discussion of quality of explanations would not disappoint me if it were provided by a good undergraduate or graduate student.

The third criterion for the best explanation that ChatGPT listed is predictive power, in agreement with many philosophers and scientists. On my own view, predictions are just another contribution to explanatory power (Thagard 1992), but ChatGPT describes 4 theories that gained plausibility by making successful predictions: general relativity, quantum mechanics, plate tectonics, and the bacterial theory of ulcers. These descriptions suggest a strong grasp of why predictive power has often been taken as relevant to theory evaluation.

Philosophers have analyzed simplicity in different ways, but ChatGPT advocates the one I prefer: making as few assumptions as possible. It accurately describes how simplicity in this sense contributed to acceptance of Copernican heliocentrism, Newton's laws of motion, Darwin's theory by natural selection, and Mendelian genetics. I would have commended an advanced graduate student for coming up with such a compelling account.

The criterion of consistency with established theories is tricky to apply because science sometimes has revolutions which require the rejection of previously accepted views (Kuhn 1970, Thagard 1992). Overall, however, we can expect theories in a domain to be compatible with accepted theories in related domains. One of the reasons for rejecting astrology and ESP is that they contradict theories that are legitimately believed in physics and psychology. ChatGPT provides a reasonable account of how partial consistency is relevant to the acceptance of relativity theory, quantum mechanics, Mendelian genetics, and the germ theory of disease.

Ignoring falsifiability and the vague idea of scope, and taking into account the many evaluations that I mentioned concerning domains, I would say that ChatGPT has a highly sophisticated understanding and practice of evaluative explanatory coherence. In most cases, it would do inference to the best explanation better than \people, because it is not subject to the many biases and fallacies to which humans succumb. People frequently engage in motivated reasoning in which they reach conclusions based on their personal goals rather than the evidence (Kunda 1990, Thagard 2024). ChatGPT gives an excellent analysis of the drawbacks of motivated reasoning for evaluating explanatory hypotheses, along with suggestions for better procedures based on empirical evidence. It similarly recognizes why *post hoc ergo propter hoc* (also known as the fallacy of false cause) is problematic and proposes superior alternatives.

After a week's delay, I again gave ChatGPT the prompt: "If one hypothesis H1 explains 3 facts, and another H2 explains 2 facts, which is better?" The answer was highly similar to the first answer, but not exactly the same because ChatGPT employs a probabilistic rather than deterministic selection of the next word to produce. This selection is not absolutely random, but rather is chosen from among the most highly rated selections. I have no idea why the second answer omitted falsifiability. Conveniently, it added a practical example from science, presumably because my earlier queries that day had frequently asked for historical examples. It volunteered an excellent overall assessment of the superiority of Copernican over Ptolemaic models of the solar system.

ChatGPT's description of how it makes this assessment is rather vague, mentioning its usual methods of machine learning and information retrieval which enable it to incorporate multiple criteria, weighting and synthesis of information, and feedback from users. When I asked it to compare its approach to my computational theory of explanatory coherence, ChatGPT correctly described my account along with relevant similarities and differences to its own approach. Similarly, ChatGPT does a fine comparison of its approach with Bayesian probability theory, and even does an amazingly deep assessment of the strengths and limitations of explanatory coherence versus Bayesian probability! It recommends that Bayesian or explanatory coherence might be preferable in different domains according to the dynamics of the field of application, the type of data available, and the purpose of analysis. This comparison sounds to me like the original work of an advanced researcher, potentially worthy of publication in a good journal in philosophy of science or cognitive science. Generative AI now provides a third general computational approach to evaluating competing explanations, capable of synthesizing explanatory coherence and Bayesian approaches. My overall assessment is that ChatGPT is excellent at evaluative explanatory inference.

## Conceptual Issues

My tests of the efficacy of ChatGPT to perform explanatory inference have presupposed that it is capable of explanation and inference. Consider the following challenge:

> Thagard, you are so gullible! You've been hoodwinked by the apparent linguistic fluency of ChatGPT to think that it actually understands what it's doing. You are completely misguided in supposing it can do explanatory inference, because it has no clue about explanation, understanding, meaning, causality, or creativity. The program may be able to fake explanatory inferences, but it doesn't actually make any. The answer to whether ChatGPT does explanatory inferences of any kind is a flat *no*.

I will respond to these accusations systematically. ChatGPT is still limited in some of these respects compared to humans, but the limitations do not undermine the claim that ChatGPT and similar models are capable of explanations.

### Explanation

In an opinion piece in the New York Times, the eminent linguist Noam Chomsky and his colleagues argue emphatically that ChatGPT and its ilk operate with a fundamentally flawed conception of language and knowledge. They claim that their reliance on machine learning and pattern recognition makes them incapable of explanation (Chomsky, Roberts, and Watumull 2023):

> Such programs are stuck in a prehuman or nonhuman phase of cognitive evolution. Their deepest flaw is the absence of the most critical capacity of any intelligence: to say not only what is the case, what was the case and what will be the case — that's description and prediction — but also what is not the case and what could and could not be the case. Those are the ingredients of explanation, the mark of true intelligence.
>
> Here's an example. Suppose you are holding an apple in your hand. Now you let the apple go. You observe the result and say, "The apple falls." That is a description. A prediction might have been the statement "The apple will fall if I open my hand." Both are valuable, and both can be correct. But an explanation is something more: It includes not only descriptions and predictions but also counterfactual conjectures like "Any such object would fall," plus the additional clause "because of the force of gravity" or "because of the curvature of space- time" or whatever. That is a causal explanation: "The apple would not have fallen but for the force of gravity." That is thinking.
>
> The crux of machine learning is description and prediction; it does not posit any causal mechanisms or physical laws.

This argument seems to be based on general ideas about machine learning, not on examination of what ChatGPT actually does. Interrogation shows that ChatGPT is highly sophisticated in its causal and counterfactual reasoning.

I asked ChatGPT 4 what happens when someone with an apple in hand opens the hand. The program responded with a 100-word paragraph that stated that the apple will fall because of the force of gravity in accord with Newton's laws of motion. When asked what would have happened if the hand not been opened, ChatGPT responded that the apple would not have fallen because the force from the hand would balance the force of gravity.

Even more impressively, ChaGPT gives a fine answer to the question of what would have happened if gravity did not exist and the hand is opened. It said that the apple would not fall because without gravity there would be no force pulling it downward. ChatGPT 3.5 gives similar but briefer answers. I put the same three questions to my son Adam, an engineer well-trained in physics, who gave similar but less-detailed answers. Accordingly, Chomsky's claims about the limitations of ChatGPT are refuted by its performance on his own example. The performance of Google's Gemini model is similar to that of ChatGPT.

ChatGPT can not only make reasonable judgments about the truth or falsity of counterfactual conditionals, it is surprisingly sophisticated about how to do so. It outlines several approaches to the difficult problem of assessing the truth of counterfactual conditionals, including possible world semantics favored by some philosophers, and causal modeling favored by some AI researchers. If you do not believe that ChatGPT is excellent at counterfactual reasoning, just query it, for example about what would have happened if the US had not dropped atomic bombs on Japan in 1945.

But does ChatGPT know what an explanation is? It provides as good a definition as can be found in dictionaries, which is not surprising because it has probably been trained on multiple electronic dictionaries. But it can also perform a richer kind of conceptual analysis based on a more psychologically realistic account of concepts as a combination of standard examples, typical features, and contributions to explanation (Blouw, Solodkin,

Thagard, and Eliasmith 2015; Thagard 2019). ChatGPT readily generates 5 good examples of explanations, 5 typical features, and 5 explanatory uses of the concept of explanation. Humans would have to think hard to do as well.

**Understanding**

But does ChatGPT understand anything? The model is remarkably modest about its capacity for understanding, proclaiming that its understanding is fundamentally different from that of humans, because it is based only on the data on which it has been trained without the personal experiences and emotions of people. Granted, understanding in people can sometimes involve a feeling such as "I've got it", but this feeling is often bogus as when people listen to politicians like Donald Trump and think they understand world politics and economics.

A more objective account views understanding as (1) connecting something coherently with what is already known, (2) applying knowledge of it in new situations, (3) being able to generalize about it, (4) thinking deeply about it, and (5) communicating this knowledge to others. ChaGPT can already do all of these. For example, I would say that ChatGPT has a good understanding of elephants, because it (1) lucidly and comprehensively describes their evolutionary origins, (2) predicts likely behaviors in unusual situations such as a supermarket, (3) generalizes about their physical and behavioral properties, (4) insightfully identifies ways in which elephants are analogous to humans along multiple dimensions, and (5) produces highly readable answers to all these questions as communications to humans.

Geoffrey Hinton contends that generative AI has a degree of understanding:

> People say, It’s just glorified autocomplete. Now, let’s analyze that. Suppose you want to be really good at predicting the next word. If you want to be really good, you have to understand what’s being said. That’s the only way. So by training something to be really good at predicting the next word, you’re actually forcing it to understand. (quoted in Rothman 2023, p. 30).

ChatGPT’s modesty about its own capacity for understanding may be based on training by humans instructed to keep it from scaring its users. I agree that current generative AI models lack emotions and consciousness, but do not see these as complete impediments to having understanding (Thagard 2021). Machines using generative AI can use alternative mechanisms to work around the cognitive benefits that humans get from focused evaluations provided by conscious emotions.

The major limitations of ChatGPT compared to human understanding reflect its current lack of interactions with the world. Humans, especially young children, come to understand the world by multiple senses and especially by acting on the world and moving objects. The imminent integration of generative AI models with robots that do interact with the world could transcend this limitation, which is also relevant to questions about causality and meaning.

**Causality**

Initially, ChatGPT seems to have a solid understanding of a cause as something that brings about an effect, with abundant examples such as that smoking causes cancer. It recognizes typical features of causal relations, including temporal precedence, covariation, and elimination of alternative factors. Causal relations contribute to explanations by identifying mechanisms, clarifying relationships, predicting outcomes, and providing control. It generates excellent examples of how causality is relevant to determining the truth or falsity of counterfactual conditionals such as "If the patient had received the vaccine, they would not have contracted the disease." ChatGPT's verbal comprehension of causality is comparable to top human causal reasoners such as epidemiologists who have developed elegant methods for determining the causes of diseases (Dammann, Poston, and Thagard 2019).

ChatGPT gives a fine verbal account of the difference between pushes and pulls with examples from many domains. But ChatGPT acknowledges that human understanding of the difference is enhanced by physical experiences, sensory feedback, and emotional states such as effort, fatigue, and motivation. The emotional and conscious aspects of pushing and pulling are beyond the capacity of ChatGPT, but robots are already capable of pushing and pulling. So AI models connected with robots should be able to learn from the robots' behaviors to identify physical correlates of pushing and pulling, but will still not have sensory experience of those actions.

The concept of causality is not amenable to strict definition, but can be analyzed in terms of exemplars, typical features, and explanations (Thagard 2019, p. 132). Standard examples include pushes, pulls, and motions, which AI models will comprehend through connections with robots. AI models already have a good verbal grasp of typical features of causality that include regularities, manipulations, interventions, statistical dependencies, and networks. Finally, ChatGPT can intelligently discuss how causes provide explanations and how causality can be explained by underlying mechanisms.

Alison Gopnik is a development psychologist famous for her research on sophisticated causal reasoning in children (Gopnik et al. 2004). She and her colleagues argue that the new AI models are excellent at imitation, but are incapable of the kind of innovation that small children can do (Yiu, Kosoy, and Gopnik 2023, Kosoy et al. 2023). The argument is based on the failure of the large language model LaMDA (produced by Google) to accomplish a well-known causal inference task. In this task, children are able to determine which objects are "blickets" on the basis of whether they set off a machine rather than on non-causal features of shape and color.

I asked ChatGPT to solve a version of the blicket detection problem based on Gopnik's original 2000 experiment (Gopnik and Sobel 2000). I replaced the term "blicket" by "gooble" so that ChatGPT could not simply look up the answer from published papers. ChatGPT instantly inferred that setting off the machine was the key feature rather than shape or color, and got the right answer about which object was a gooble.

Moreover, when asked how it reached its conclusion, ChatGPT described sophisticated causal reasoning with hypotheses about what factors might set off the machine. When queried, it reported not using Bayesian probabilities because the relevant probabilities were not available. I suspect the same is true of children.

This analysis is too subtle to have been produced through reinforcement learning by humans rather than training from examples. So I see no reason to believe that ChatGPT is merely imitative rather than innovative, especially given the many examples of creative hypothesis formation that I have described. I attribute the earlier failure of Gopnik and her colleagues to find child-level causal reasoning to their use of a now-obsolete model. Google has replaced LaMDA by Gemini, with many more parameters, and it also behaves like children on the blicket test. I predict that ChatGPT 4, Gemini, Claude 3, and Llama 3 can handle the many other causal reasoning tasks that Gopnik and her colleagues have studied in children.

One aspect of causality that ChatGPT currently lacks is a deep biological understanding of time. Like any computer program, it can precisely identify time by seconds, minutes, and dates, but biological systems such as humans lack such clocks, so how do they manage time in ways required for causal reasoning and many other functions? I think that the two key neural mechanisms are time cells in the brain that keep track of small intervals, and memory units that bind intervals with other information such as spatial location (Thagard forthcoming, Eichenbaum 2014, Voelker and Eliasmith 2018). These mechanisms allow animals to keep track of relations of before, after, and simultaneous, thereby managing the temporal precedence and covariation aspects of causal reasoning without explicit clocks. Analogs of these mechanisms could potentially be implemented in AI models, but they can do well at causal reasoning without them because of computational clocks and verbal representations of time. Although causal reasoning by generative AI models is not exactly the same as that performed by humans and other animals, it is nevertheless impressive and displays substantial understanding of causality.

**Meaning**

A radical critique of generative AI would say that these models are incapable of explanation, understanding, and causal reasoning because the sentences that they fluidly generate are meaningless. John Searle (1980) claimed on the basis of his Chinese Room thought experiment that computers have syntax but no semantics. They are like a person in a room who gets Chinese symbols as inputs and produces them as outputs by looking up rules in a table, without understanding the symbols or the rules. This analogy has many flaws that are particularly evident in the operation of the new AI models, which are far more than lookup tables: they are trained on vast amounts of data that can produce networks with more than a trillion parameters, enabling them to generate complex pieces that answer complex questions. The attention mechanism allows them to relate many symbols to each other and produce rich amounts of word-to-word meaning, i. e. the meaning that symbols get from their relations to other symbols.

What about the other main kind of meaning based on connections to the world? Searle could argue that ChatGPT symbols are not about the world because the program has had no interactions with the world. Several responses apply. First, ChatGPT does get indirect connections with the world because the texts on which it has been trained were produced by people who did observe and interact with the world. Such connections are second-hand, but so are many of the connections that people use. I have never been to India, but I have a pretty good understanding of the Taj Mahal from reading about it.

Second, ChatGPT can already take visual inputs, so its internal representations can be partly based on pictures, not just the words that operate in the Chinese Room. This possibility allows meaning in ChatGPT to be visual as well as verbal. Third, as I have frequently mentioned, the current disconnection of generative AI models from the world is temporary and will soon be overcome through robotic interactions that could potentially be tactile, auditory, and olfactory as well as verbal and visual. At that point, ChatGPT will be capable of multimodal meaning that puts the last nail in the coffin of Searle's thought experiment. The long-established operation of driverless cars already shows that machines can use sensors to learn how to operate in the world (Parisien and Thagard 2008).

**Creativity**

Peirce's abduction and what I have called explanatory inference and hypothesis formation are supposed to be major sources of creativity. But is ChatGPT capable of being creative? The key features of creativity are that a product be novel, surprising, and valuable (Boden 2004, Simonton 2004). ChatGPT is incapable of emotions like surprise, and its conception of what is valuable is limited by the fact that it has no values, which are goals associated with emotions (Thagard 2019).

However, people can judge whether the contributions of ChatGPT qualify as creative, and AI has already made contributions to research on drug discovery, protein folding, materials science, climate change, synthetic biology, and astronomy (Microsoft Research AI4Science 2023). Probably the most surprising and valuable application of generative AI to date is DeepMind's AlphaFold, which accurately predicts the 3D structure of proteins, with many biological applications (Callaway 2022). None of these are Nobel-Prize level discoveries, but most of the creativity in humans is not at that level either. A group of scientists at leading universities are collaborating to use multidisciplinary AI to tackle major problems (polymathic-ai.org).

ChatGPT and similar models also produce impressive results in other fields such as game playing, poetry, short stories, and dramatic dialogues. AI models score higher than humans on divergent thinking tasks (Hubert, Awa, and Zabelina 2024). Accordingly, I see no reason to doubt the potential creativity of ChatGPT and similar models. Conceptual issues about explanation, understanding, causality, meaning, and creativity do not undermine the capabilities of generative AI.

## Implications

My answer to the question in the title is clear: Yes, ChatGPT can make explanatory inferences, including both creative ones that form new hypotheses and evaluative ones that determine which hypotheses are part of the best explanation of the evidence. My reasons for reaching this conclusion are not selective anecdotes, but rather systematic tests on benchmarks concerning domains, modalities, hypothesis formation, and hypothesis evaluation. ChatGPT is limited in modalities to verbal and visual inferences, lacking capabilities for sound, touch, smell, taste, internal sensations, and emotions. But its performance on other benchmarks is comparable to highly intelligent humans. Arguments that such models are inherently limited in their capacities for explanation, understanding, causality, meaning, and creativity are easily rebutted.

The realm of explanatory reasoners has recently expanded to include computers as well as people. But the giant remaining question is: How did this expansion happen? The new AI models, based on the Transformer technology using vector mathematics, attention, and neural networks trained by backpropagation, were trained only to predict the next word in a sequence. They were not trained to generate or evaluate explanatory hypotheses, so how did they get so good at it? ChatGPT and Gemini answer this question by pointing to their vast training data, advanced architectures using deep learning and attention, contextual understanding, pattern recognition, fine tuning, and capacity to generate novel interpretations. But how does this all add up to remarkable approximations to human intelligence?

Arora and Goyal (2023) suggest that the power of generative AI comes from the emergence of complex skills. Large language models develop skills for which they were not explicitly trained. For example, pronoun resolution is crucial for language understanding in sentences like “Bob punched Doug because he insulted him.” ChatGPT can use background knowledge about insults to infer that “he” most likely refers to Doug and “him” refers to Bob. The skill of doing pronoun resolution is an emergent property of the trained neural networks. More complex skills such as putting together a coherent paragraph or short story emerge from the interaction of simpler skills.

Arora and Goyal do not define emergence, but the usual account in complex systems applies: a property is emergent if it belongs to a whole but not to its parts or the sum of its parts, because it results from the interactions of the parts (Bunge 2003, McClelland 2019, Thagard 2019, Wimsatt 2007). Generative AI seems to be capable of recursive emergence, where emergent processes combine to produce further novelty; I have used recursive emergence to explain the origins of human consciousness (Thagard forthcoming). Such emergence of complex skills shows why the new AI models are much more than “stochastic parrots” that merely use statistics to imitate humans.

But what makes the new AI models so good at developing emergent skills? Geoff Hinton makes the interesting suggestion that the key is compression of all the information in training sources into a trillion or so parameters (Diamandis 2024). He suggests that such compression requires the resulting networks to recognize similarities and generalize, in ways that then make them capable of complex reasoning. I hope that further investigations and development of future models will provide deeper explanations of how this works. AI researchers have created a form of alien intelligence that we are only beginning to understand.

At the basic level of program operation, creative and evaluative explanatory inferences in ChatGPT are the result of training of neural networks using backpropagation, attention and other mechanisms. But they can also be viewed as separate skills that emerged from other simpler skills produced by training over large data bases. We do not need to know exactly why the new AI models work so well to appreciate their accomplishments with respect to explanatory inference. Charles Peirce, Lorenzo Magnani, and other classic researchers on abduction should be happy to have new subjects and tools for future investigations.

**Acknowledgements:** For helpful suggestions, I am grateful to Chris Eliasmith, Keith Holyoak, Jay McClelland, Lorenzo Magnani, Dean Simonton, and 2 anonymous referees.